\documentclass[conf]{new-aiaa}

\usepackage[utf8]{inputenc}
\usepackage{graphicx}
\usepackage{amsfonts,amsmath}
\usepackage{siunitx}
\usepackage{longtable,tabularx}
\usepackage{adjustbox}
\usepackage{dsfont}
\usepackage{subcaption}
\usepackage{comment}
\usepackage{svg}
\usepackage{gensymb}
\usepackage{multirow}
\usepackage{cleveref}
\usepackage{array}
\newcolumntype{L}[1]{>{\raggedright\arraybackslash}m{#1}}

\newcommand\norm[1]{\left\lVert#1\right\rVert}

\usepackage{soul}

\usepackage{xcolor}
\definecolor{rev1}{HTML}{FF999A} 
\definecolor{rev2}{HTML}{F3F298} 
\definecolor{rev3}{HTML}{B2E0AE} 
\definecolor{other}{HTML}{C8C7FF} 
\sethlcolor{rev1}

\soulregister\cite7
\soulregister\citep7
\soulregister\citet7
\soulregister\ref7
\soulregister\eqref7

\title{A Neural Envelope Method for Rapid Gradient-based Optimum Takeoff Trajectory Design}

\author{Samuel Sisk\footnote{Doctoral Student. Department of Mechanical and Aerospace Engineering, AIAA Student Member.}}
\author{Xiaosong Du\footnote{Assistant Professor. Department of Mechanical and Aerospace Engineering, AIAA Member}} 
\affil{Missouri University of Science and Technology, Rolla, MO 65409, USA}

\begin{document}

\maketitle


\begin{abstract}
Electric vertical takeoff and landing (eVTOL) aircraft emerge as a new transportation option for mitigating ground traffic congestion. 
Multidisciplinary analysis and optimization (MDAO) has been introduced to reduce the energy consumption during takeoff, thereby expanding the range and endurance.
Surrogate-based MDAO lowers the computational burden of optimization,
but multimodality makes both simulation-based and surrogate-based optimizations challenging to identify global optimal design. 
Convex envelope converts multimodal spaces into unimodal spaces, permitting efficient convex optimization; however, convexifying a design space may result in physically infeasible "optimal" designs due to complex design constraints. 
To address the above challenges, this work proposes the coupling between physics-constrained generative adversarial networks (physicsGAN) and neural envelope to transform original, constrained, non-convex optimizations into unconstrained, convex optimizations. 
Specifically, the physicsGAN discovers a feasible design space where all design candidates inherently satisfy all constraints, by penalizing surrogate-predicted constraint violations during training.
Moreover, the neural envelope constructs a neural network-based convex envelope for the transformed objective function within the physicsGAN-discovered feasible space. 
Note that the neural envelope only needs to deal with objective function since constraints are inherently satisfied in the feasible space. 
The neural envelope is demonstrated on takeoff trajectory optimizations of the Airbus $A^3$ Vahana eVTOL aircraft. 
Results show that neural envelopes achieve a 99.91\% mean accuracy within a continuous flight-condition space compared with reference convex envelopes.
Neural-envelope-enabled optimizations achieve 99.84\% accuracy on the global optimum design, costing only 0.04 seconds on average.
The neural envelope accelerates the trajectory optimization by 53 and 162 times on average compared to the physicsGAN-enabled gradient-based and global optimizations, respectively.
\end{abstract}

\section{Introduction}
\label{sec:introduction}
Urban mobility delays caused by congestion in major United States cities rose by 20--40\% from 1982 to 2017~\cite{texas2019transportation}.
In response to this mounting traffic load, urban air mobility (UAM) has arisen as an additional regional transportation mode meant to increase network capacity; projections foresee 4,000 aircraft carrying 80,000 daily passengers in a fast-growing two-billion-dollar market~\cite{goyal2018urban}.
UAM enables point-to-point movement of passengers and cargo through decentralized vertiport networks spread over metropolitan regions~\cite{straubinger2020overview}, and optimizing their layout can reduce the associated infrastructure costs~\cite{vascik2019development}.
However, today's UAM depends largely on conventional helicopters, which results in limited market penetration and high costs per passenger~\cite{vascik2017systems}.

Electric vertical takeoff and landing (eVTOL) aircraft represents a promising next step for current UAM operations, since they substantially reduce seat-mile and total operating costs compared with traditional helicopter air taxis~\cite{duffy2017study}.
In addition, geometric programming has shown that electric short takeoff and landing flights in cities are viable~\cite{courtin2018feasibility}, and life cycle assessments suggest that eVTOL aircraft powered by low-carbon grids has lower lifetime emission than conventional cars~\cite{andre2019robust}.
Motivated by such economic and environmental advantages, leading aviation companies, including Ehang~\cite{EHang} and Joby~\cite{Joby}, regard eVTOL aircraft as profitable transport options~\cite{goyal2018urban}.
Despite these advantages, broad adoption of eVTOL aircraft is still hindered by open issues, such as energy consumption, thermal management, vibration, and noise pollution during operations.

These limitations have been investigated in many studies~\cite{edwards2020evtol,falck2017trajectory,yang2021challenges,sripad2021promise,harrison2019using}, which have identified mechanical and operational strategies for mitigating excessive noise and vibration.
\citet{zhang2024high} used high-fidelity aeroacoustic modeling to tie increased propeller noise to rotor phase synchronization as well as aerodynamic interference among the fuselage, wings, and propellers.
Trajectory optimization manages to simultaneously balance noise emissions, energy use, and flight volume~\cite{gao2024noise}.
\citet{falck2017trajectory} targeted energy and thermal limits and showed that trajectory planning under thermal constraints reduced both motor size and thermal loads.
Moreover, \citet{chauhan2020tilt} employed multidisciplinary analysis and optimization (MDAO) for energy-efficient eVTOL aircraft takeoff paths and showed that acceleration constraints raised energy requirements considerably while stall and wing loading had negligible effects.
However, conventional MDAO repeatedly evaluates physics-based simulation models, prohibiting rapid decision-making. 

Replacing costly high-fidelity simulations with surrogate models paves an avenue for efficient, or even real-time, design optimization~\cite{mlaso, Peherstorfer2018a, Leifsson2011surrogate}.
\citet{alba2017surrogate} reduced the number of function evaluations needed to assess wing-propeller interactions by half through surrogate-based MDAO.
Other machine learning strategies, such as reduced-order models and deep neural networks (DNN), are widely to emulate complex engineering models.
For instance, the reduced-order models based on proper orthogonal decomposition by \citet{swischuk2020learning} approximated nonlinear partial differential equations in combustion. 
\citet{lazzara2022surrogate} combined long short-term memory (LSTM) and feedforward DNN to predict dynamic landing loads in aircraft structural design.
Direct surrogate modeling could incur excessive training cost, leading to the studies of dimensionality reduction. 

Dimensionality reduction strategies, such as proper orthogonal decomposition and autoencoder, considerably improve surrogate modeling in high-dimensional engineering applications~\cite{Bird2021,Hou2022,OLEARYROSEBERRY2022115730}.
Generative artificial intelligence leads a new trend for dimensionality reduction~\cite{du2026generative}.
Generative adversarial networks (GAN)~\cite{Goodfellow2014}, originally developed to reproduce the distribution of existing data, inherently provide implicit design space reduction by generating plausible design geometries, such as smooth airfoils~\cite{Du2021a} and takeoff control profiles~\cite{sisk2024surrogate}.
\citet{sisk2024surrogate} introduced a twin-generator GAN (twinGAN) for eVTOL aircraft trajectory optimization that reduced dimensionality in two ways: generating only near-optimal takeoff paths shrinks the space implicitly, and the original 41-dimensional design space is explicitly compressed to just four dimensions.

Moreover, optimizations based on simulations or surrogates often suffer from complex, nonlinear design constraints in practical engineering applications~\cite{mdobook}.
Conventional constraint-handling approaches frequently prove inadequate; penalty methods are easy to implement yet have trouble guaranteeing feasibility at the optimum~\cite{mdobook}, whereas advanced algorithms such as sequential quadratic programming and interior-point methods require considerable mathematical expertise to set up properly~\cite{Karmarkar1984,Gill2002a}.
More recently, generative artificial intelligence has offered an attractive alternative that builds real-world constraints directly into the training of diffusion models~\cite{power2023sampling,zhu2024controltraj}, transformers~\cite{guffanti2024transformers,johnson2024zero}, and GAN~\cite{yang2020physics,yu2024physics}.
Existing literatures have succeeded in producing constrained outputs---for instance by training GANs on feasible datasets to create new chemicals with 80--90\% feasibility~\cite{dan2020generative} or by incorporating discrete logical constraints into adversarial networks~\cite{di2020efficient}---but these earlier methods cannot rigorously enforce or systematically comply with constraints.
\citet{sisk2026physics} developed the physics-constrained GAN (physicsGAN), which learned continuous physical constraints to produce a feasible design space enabling unconstrained optimization on an originally constrained problem.

Furthermore, simulation- and surrogate-based optimizations also struggle with multimodality, which results in higher complexity search for global optimum.
In a pooling optimization problem, \citet{pham2009convex} applied the convex-hull method and found that an optimization framework built on a convex envelope improved efficiency by an order of magnitude across the board.
\citet{yang2021optimization} performed control based on convex optimization using input-convex neural networks, and stressed the value of reliably guaranteeing that the global optimum is reached.
A nonlinear optimization problem was linearized and convexified with a max-affine network by \citet{ahmadi2013piecewise}, whose max-affine-based optimizations were two orders of magnitude faster than the reference optimization.
Nevertheless, these neural-network-based approaches show that convexification methods may be limited in representational capacity.
Besides, convexification-based optimum design could be physically infeasible. 

To expand representational capacity of convex envelope in high-dimensional design space and guarantee feasible solutions, this work proposes the combination of neural envelope and physicsGAN.
The neural envelope is trained on convex envelopes of feasible data produced by physicsGAN~\cite{sisk2026physics}, allowing it to approximate the convex envelope over the whole feasible design space.
The physicsGAN is trained on a feasible dataset produced by twinGAN and surrogates and penalized on constraint violations according to surrogate predictions during training. 
The neural envelope then learns the convex envelope of the physicsGAN; this cuts the number of model calls, gives a unimodal optimization landscape, and guarantees that feasible global optimum is found.
Please refer to Secs~\ref{sec:physicsGAN} and \ref{sec:envelope} for details. 

The rest of the article is organized as follows.
Section~\ref{sec:background} presents the underlying simulation frameworks, the data-driven twinGAN, the necessary surrogate models, and the physicsGAN.
Section~\ref{sec:envelope} describes the process used to convexify the physicsGAN in detail.
Section~\ref{sec:results} applies the proposed method to optimal eVTOL takeoff trajectory design and compares its performance with both conventional simulation-based MDAO and twinGAN-enabled surrogate-based MDAO.
Lastly, Section~\ref{sec:conclusion} recaps the main contributions of this work and suggests directions for future research.

\section{Background}
\label{sec:background}
This section introduces the simulation models, twinGAN, and surrogate models used in whis work, followed by elaborations on physicsGAN.

\subsection{Aircraft and Simulation Models}
\label{sec:Sim Model}

\begin{figure}[htbp!]
    \centering
    \label{fig:simulation}
    \begin{subfigure}{.5\textwidth}
        \centering
        \includegraphics[scale=0.7]{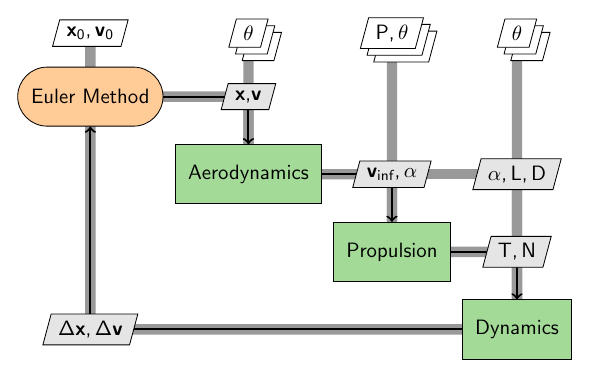}
        \caption{}
        \label{fig:sim_model}
    \end{subfigure}%
    \begin{subfigure}{.5\textwidth}
        \centering
        \includegraphics[scale=0.6]{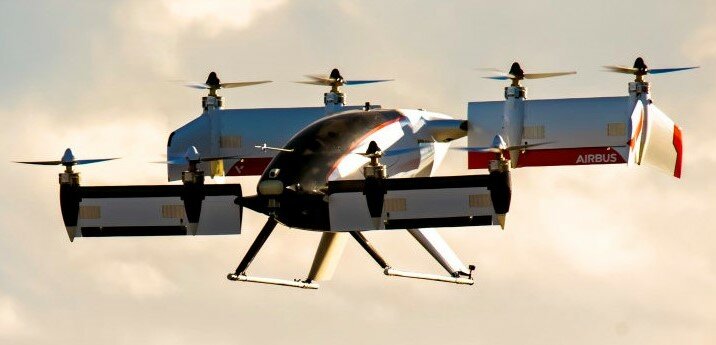}
        \caption{}
        \label{fig:Vahana}
    \end{subfigure}%
    \caption{
    The eVTOL models used in this work: (a) the flow of information through the simulation models is shown in the process diagram; (b) the Airbus $A^3$ Vahana eVTOL aircraft.}
\end{figure}

\begin{figure}[htbp!]
    \centering
    \includegraphics[width=0.8\linewidth]{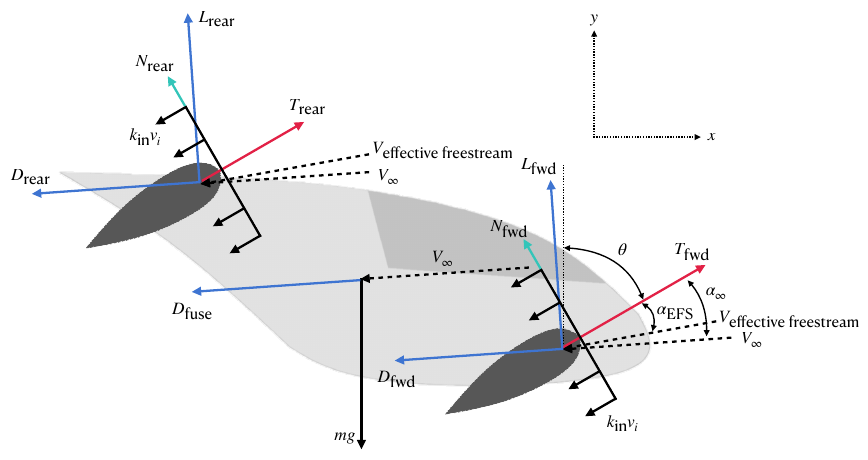}
    \caption{Dynamics model of the Airbus $A^3$ Vahana eVTOL aircraft.}
    \label{fig:Dynamics Model}
\end{figure}

This work adopts first-principle and empirical simulation models (Fig.~\ref{fig:sim_model}) to represent the Airbus $A^3$ Vahana (Fig.~\ref{fig:Vahana}) eVTOL aircraft.
Each subsystem of the simulation model determines the forces (Fig.~\ref{fig:Dynamics Model}) that act on the aircraft; the aerodynamics subsystem, for instance, yields lift and drag, while the propulsion subsystem yields the propeller-induced thrust and normal forces.
The analysis is reduced to a two degree of freedom point-mass trajectory optimization through simplifying assumptions.
A forward Euler method propagates the trajectory, which is then optimized with \texttt{SNOPT}~\cite{Gill2002a}.
Flow augmentation is omitted here because~\citet{chauhan2020tilt} concluded that its effects are negligible.
Further detail on the simulation models can be found in \citet{chauhan2020tilt}.


\subsection{Twin-Generator Generative Adversarial Networks}
\label{sec:TwinGAN}
GAN, together with other generative models, aims to produce data that closely mimics the statistical structure of a given training dataset~\cite{Goodfellow2014}.
This work selects GAN due to the integrated features of generative quality, fast inference, and two-fold dimensionality reduction~\cite{du2026generative}.  
Two neural networks compete within the GAN architecture: a generator, which maps random noise ($\mathbf{z}$) to synthetic data ($\mathbf{g}$) imitating the real training data ($\mathbf{x}$); 
and a discriminator, which differentiates the generated samples from the training data.
The following minimax loss function mathematically expresses the adversarial mechanism between the generator and the discriminator:
\begin{equation}
    \label{eq:GAN}
    \min_{G} \max_{D} V\left(D,G\right)=E_{\mathbf{x} \sim \mathbf{p}_{\text{data}}\left(\mathbf{x}\right)}\left[\log D\left(\mathbf{x}\right)\right]+E_{\mathbf{z} \sim \mathbf{p}_\mathbf{z}\left(\mathbf{z}\right)}\left[\log \left(1-D\left(G\left(\mathbf{z}\right)\right)\right)\right],
\end{equation}
where $V$ is the loss function capturing the interaction between the generator $G$ and discriminator $D$, $\mathbf{x}$ is drawn from the true data distribution $\mathbf{p}_{\text{data}}$, and $\mathbf{z}$ is drawn from a noise distribution $\mathbf{p}_\mathbf{z}$ (Uniform(0, 1) distributions in this work).
In training, the discriminator aims to assign a probability of 1 to real data ($p_\mathbf{x}$) and 0 to synthetic data ($p_\mathbf{g}$), whereas the generator tends to compete by driving $p_\mathbf{g}$ toward 1; this continues until the synthetic samples $\mathbf{g}$ cannot be differentiated apart from the real data $\mathbf{x}$.

\begin{figure}[!h]
    \centering
    \includegraphics[scale=0.75]{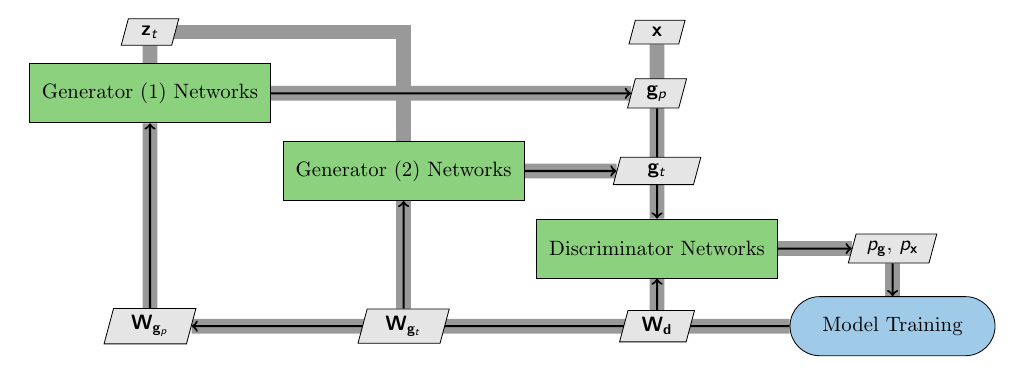}
    \caption{Architectural overview of the twinGAN model.}
    \label{fig:TwinGAN_XDSM}
\end{figure}

This work uses the twinGAN model~\cite{sisk2024surrogate} consisting of two generators (Fig.~\ref{fig:TwinGAN_XDSM}), which share the same noise inputs ($\mathbf{z}_t$) yet synthesize the wing angle and power profiles independently for higher flexibility.
By learning to distinguish between he reference training data ($\mathbf{x}$) and the synthesized power ($\mathbf{g}_1$) and wing angle ($\mathbf{g}_2$) control points, the twinGAN training adjusts the corresponding network weights ($\mathbf{W}_{\mathbf{g}_1}$, $\mathbf{W}_{\mathbf{g}_2}$, and $\mathbf{W}_\mathbf{d}$) using the predicted probabilities ($p_\mathbf{g}$ and $p_\mathbf{x}$).
The twinGAN's minimax loss is written correspondingly as:
\begin{equation}
    \label{eq:TwinGAN}
    \min_{G_p,G_t} \max_{D} V\left(D,G_p,G_t\right)=E_{\mathbf{x} \sim \mathbf{p}_{\text{data}}\left(\mathbf{x}\right)}\left[\log D\left(\mathbf{x}\right)\right]+E_{\mathbf{z}_t \sim \mathbf{p}_{\mathbf{z}_t}\left(\mathbf{z}_t\right)}\left[\log \left(1-D\left(G_p\left(\mathbf{z}_t\right),G_t(\mathbf{z}_t)\right)\right)\right],
\end{equation}
where $G_p$ and $G_t$ are the generators that predict the B-spline control points of the power and wing angle profiles, respectively, and $\mathbf{z}_t$ denotes the shared twinGAN noise vectors sampled from the Uniform(0, 1) distributions $\mathbf{p}_\mathbf{z_t}$.

Specifically, $G_p$ and $G_t$ each produces a separate set of 20 B-spline control points that parameterize the power and wing angle profiles, respectively.
Each generator has three hidden layers of 256, 512, and 1,024 neurons, and each layer applies a \texttt{LeakyReLU} activation ($\alpha = 0.2$) along with batch normalization using a momentum of 0.8.
A \texttt{sigmoid} activation at the generator output layers keeps the outputs consistent with the [0, 1] scaling of the reference training data.
The discriminator, in contrast, takes all 40 control points concatenated together, real or synthesized, and passes them through two hidden layers of 512 and 256 neurons with \texttt{LeakyReLU} ($\alpha=0.2$) activations, finally returning a single probability through a \texttt{sigmoid} function.

In this study, the twinGAN is trained on 1,099 optimal flight trajectories generated by varying aircraft mass ($m$) and propulsive efficiency ($\eta$).
The mass is varied by $\pm 15\%$ about a 725~kg baseline, and the propulsive efficiency by $\pm 10\%$ about a 90\% baseline.
Training runs for $3\times10^4$ epochs with a batch size of 20 samples, using the \texttt{Tensorflow}~\cite{tensorflow2015-whitepaper} implementation of the \texttt{Adam} optimizer~\cite{kingma2017adam} with a learning rate of $2\times10^{-4}$ and $\beta_1 = 0.5$.
The final model is selected by comparing the generated outputs with the reference data and picking the iteration with the smallest sum of generator and discriminator losses.

\subsection{Surrogates}
\label{sec:Surrogates}
\subsubsection{Feedforward Neural Networks}
\label{sec:fnn}
\begin{figure}[ht]
    \centering
    \includegraphics[width=\linewidth]{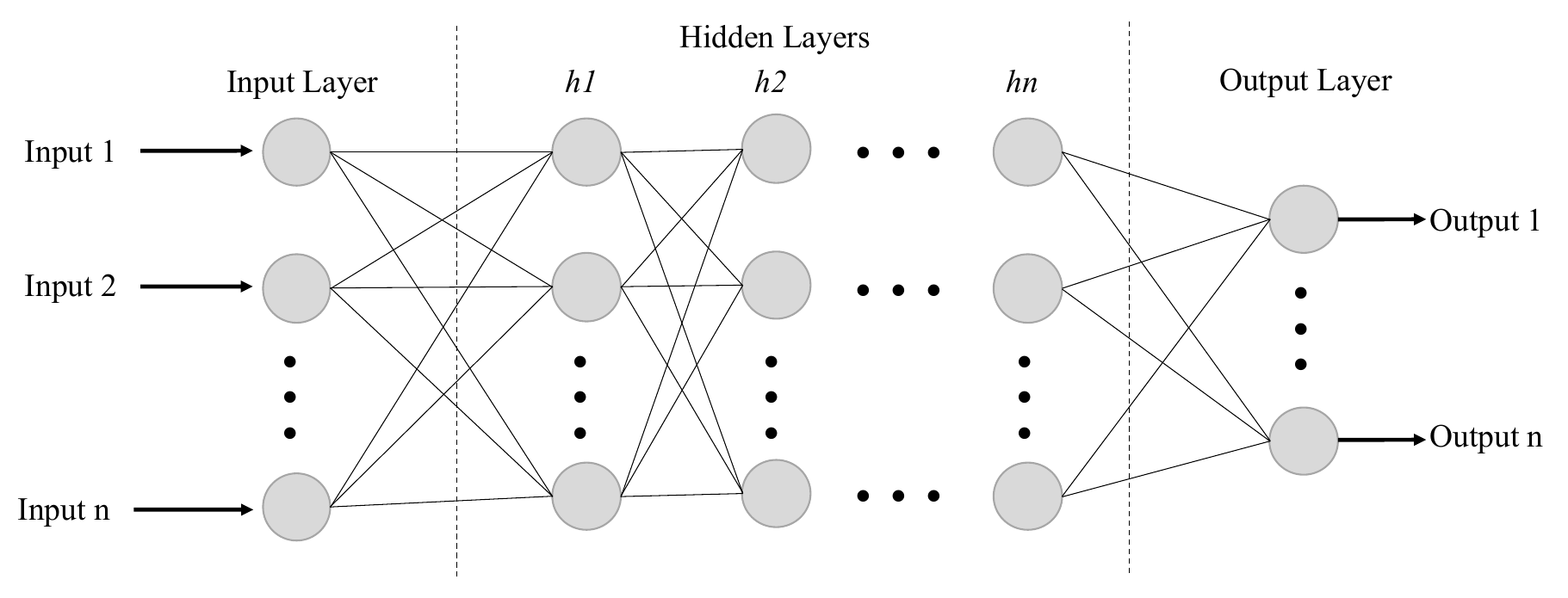}
    \caption{The general architecture of feedforward DNN.}
    \label{fig:General Deep Neural Network}
\end{figure}

DNN~\cite{werbos1982applications} is a type of machine learning inspired by neuroscience and built to learn input-output relationships from training data.
In general, DNN consists of an input layer, an output layer, and a number of hidden layers in between, with each holding multiple neurons (Fig.~\ref{fig:General Deep Neural Network}).
Every neuron in the hidden and output layers applies a linear operation to the previous layer's data and then passes the result through an activation function, which injects nonlinearity:
\begin{equation}
\label{eq:DNN}
    f(\mathbf{x}) = \phi \left( b + \mathbf{w}^\intercal \mathbf{x} \right),
\end{equation}
where
\begin{equation}
    \mathbf{w}^\intercal =
    \begin{bmatrix}
        w_1 & w_2 & \dots & w_n \\
    \end{bmatrix},
\end{equation}
\begin{equation}
    \mathbf{x}^\intercal = 
    \begin{bmatrix}
        x_1 & x_2 & \dots & x_n \\
    \end{bmatrix},
\end{equation}
$f$ is a single DNN neuron, $\mathbf{x}$ is its input vector, $\mathbf{w}$ holds the connection weights, $b$ is the bias, and $\phi$ is the activation function.
The model is optimized by repeatedly updating the weights and biases to reduce a training loss function, using gradient-based algorithms which obtain the gradients through backpropagation.

In this work, a feedforward DNN serves as the surrogate for takeoff energy consumption, which is the objective function of the takeoff trajectory optimization.
Bayesian optimization with \texttt{Keras Tuner}~\cite{omalley2019kerastuner} is used to tune the network architecture and learning rate.
The resulting DNN architecture has five hidden layers of 410, 290, 500, 310, and 500 neurons, in that order.
A \texttt{ReLU} activation follows each hidden layer, and the output layer uses a \texttt{linear} activation.
The feedforward DNN is trained for 500 epochs with a batch size of 100 to minimize a \texttt{Mean Squared Error} (MSE) training loss using the \texttt{Adam} optimizer (learning rate of $3.6\times10^{-3}$).
To monitor training, 20\% of each data batch is held out for validation, and the DNN with the lowest validation error is chosen for the surrogate-based MDAO.

\subsubsection{Long Short-Term Memory Networks}
\begin{figure}[htbp!]
    \centering
    \includegraphics[scale=0.25]{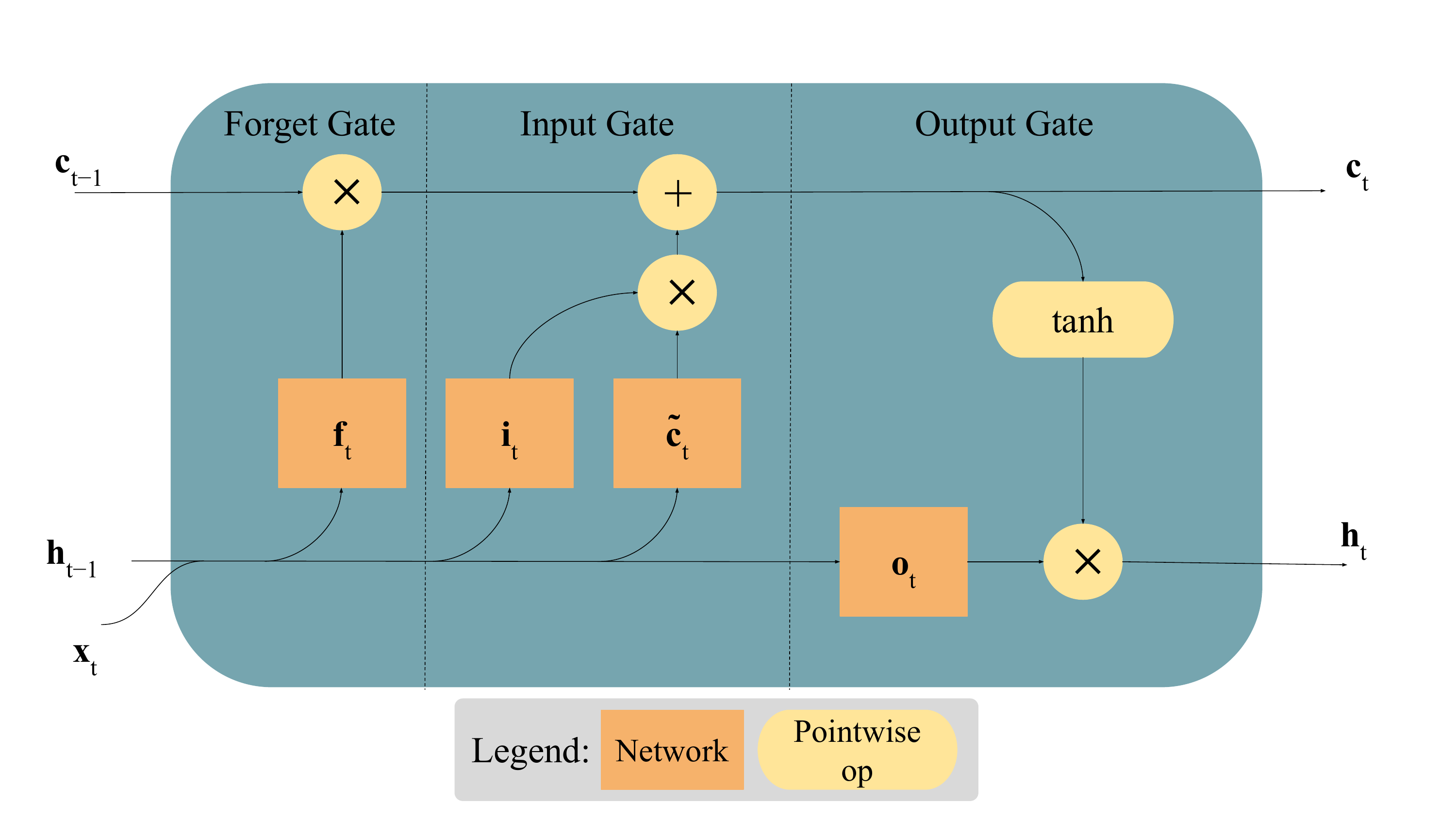}
    \caption{LSTM cells differ from feedforward DNN neurons due to their internal memory state and gates, which allow them to excel at time-sequence prediction.}
    \label{fig:LSTMcell}
\end{figure}

The feedforward DNN predicts the scalar optimization objective, but the physical constraints depend on time-series aircraft states, such as acceleration profiles.
RNN excels in handling sequential data but suffers from vanishing or exploding gradients across long sequences.
LSTM extends RNN to address the long-term dependency~\cite{Sherstinsky2020}.
In particular, LSTM uses specialized cells (Fig.~\ref{fig:LSTMcell}) with internal memory states ($\mathbf{c}$) to handle time-series data.
Additionally, the internal gating mechanism~\cite{hochreiter1997long} regulates the flow of information: the forget ($\mathbf{f}_t$), input ($\mathbf{i}_t$), and output ($\mathbf{o}_t$) gates, which decide what is discarded from, stored in, or emitted from the memory state, respectively.
\begin{equation}
    \label{eq:ForgetGate}
    \mathbf{f}_t = \sigma \left(\mathbf{W}_f(\mathbf{h}_{t-1}\frown\mathbf{x}_t)+\mathbf{b}_f \right),
\end{equation}
\begin{equation}
    \label{eq:InputGate}
    \mathbf{i}_t = \sigma \left(\mathbf{W}_i(\mathbf{h}_{t-1}\frown\mathbf{x}_t)+\mathbf{b}_i \right),
\end{equation}
\begin{equation}
    \label{eq:OutputGate}
    \mathbf{o}_t = \sigma \left(\mathbf{W}_o(\mathbf{h}_{t-1}\frown\mathbf{x}_t)+\mathbf{b}_o \right),
\end{equation}
where $\sigma$ is the \texttt{sigmoid} activation, $\mathbf{W}$ and $\mathbf{b}$ are the weight matrices and bias vectors of the corresponding gates, $\frown$ is the concatenation operator, $\mathbf{x}_t$ is the current input, and $\mathbf{h}_{t-1}$ is the cell output from the previous time step.
The memory cell is then updated as follows:
\begin{equation}
    \label{eq:CellState}
    \Tilde{\mathbf{c}}_t = \tanh{(\mathbf{W}_c(\mathbf{h}_{t-1}\frown\mathbf{x}_t)+\mathbf{b}_c)},
\end{equation}
\begin{equation}
    \label{eq:CandidateCellState}
    \mathbf{c}_t = \mathbf{f}_t \odot \mathbf{c}_{t-1} + \mathbf{i}_t \odot \Tilde{\mathbf{c}}_t,
\end{equation}
\begin{equation}
    \label{eq:CellOutput}
    \mathbf{h}_t = \mathbf{o}_t \odot \tanh{(\mathbf{c}_t)},
\end{equation}
where $\Tilde{\mathbf{c}}_t$ is the proposed candidate state, $\mathbf{c}_t$ is the updated cell memory, $\mathbf{h}_t$ is the current cell output, and $\odot$ is element-wise (Hadamard) multiplication.

In this work, LSTM surrogates predict the time-dependent state variables that formulate the constraints in takeoff trajectory optimization.
A fixed architecture is used, with four hidden layers of 200 LSTM cells each, followed by a fully connected output layer of 500 neurons, matching the number of time steps in the trajectory profiles.
As with the DNN, the hidden layers apply \texttt{ReLU} activations and the output layer a \texttt{linear} activation, and training uses the \texttt{Adam} optimizer (learning rate of $1\times10^{-3}$) to minimize the MSE loss.
This setup performs a one-to-one sequence mapping, converting a single input array (made up of $\mathbf{z}_t$, $\eta$, $m$, and $T_{\text{total}}$) into a full output time series.
Training uses a batch size of 200 with a 20\% validation split, and the model with the lowest validation loss is kept for the surrogate-based MDAO.

\subsection{Physics-Constrained Generative Adversarial Networks}
\label{sec:physicsGAN}
For physicsGAN modeling, the twinGAN is trained first to produce near-optimal control profiles, which has been proved to possess the two-fold dimensionality reduction (Sec.~\ref{sec:TwinGAN}).
The reduced latent space facilitates surrogate modeling by lowering the training cost. 
The constraint-adherence predictions from these surrogates then guide the training of the physicsGAN.
After this first phase, the twinGAN's weights must stay frozen, since changing them would change the latent space and invalidate the trained surrogates.

\begin{figure}[htbp!]
    \centering
    \includegraphics[width=\linewidth]{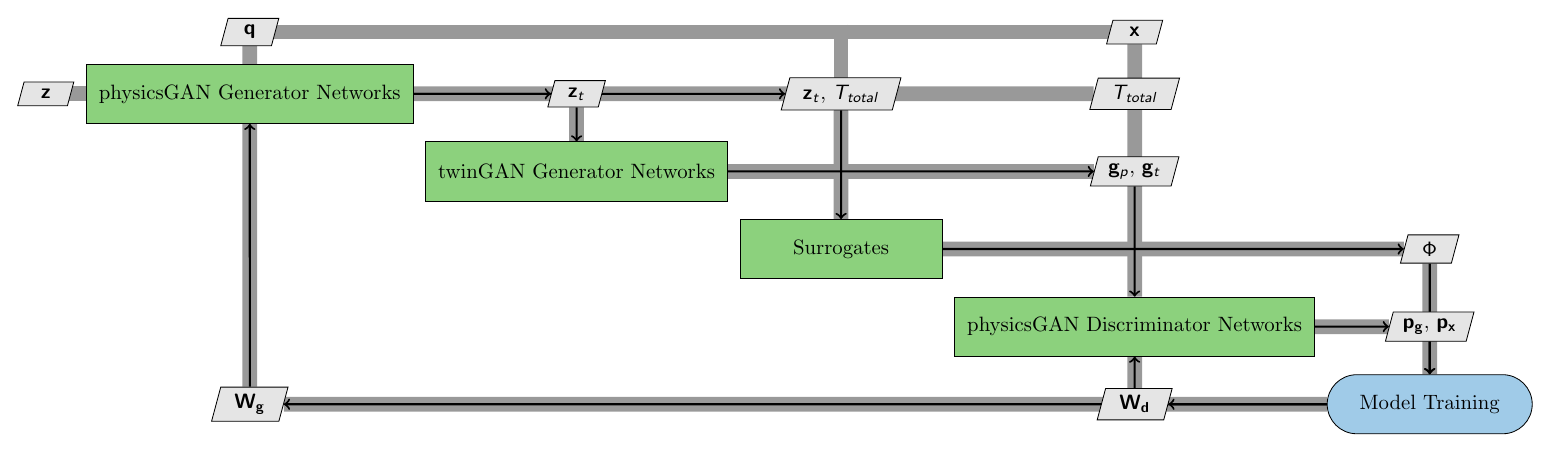}
    \caption{
    The physicsGAN architecture shows the interaction between the physicsGAN, the twinGAN generator networks, and the surrogates.
    }
    \label{fig:ConGAN_XDSM}
\end{figure}

The overall structure and workflow of physicsGAN are shown in Fig.~\ref{fig:ConGAN_XDSM}.
The physicsGAN generator receives a separate set of noise variables ($\mathbf{z}$) together with the flight conditions ($\mathbf{q}$) and returns twinGAN noise variables ($\mathbf{z}_t$) and flight duration (${T}_{\text{total}}$).
In this way, the physicsGAN discovers the feasible subset of the twinGAN's distribution of realistic trajectories.
From these outputs, the frozen twinGAN generators synthesize the matching power ($\mathbf{g}_p$) and wing angle ($\mathbf{g}_t$) profiles.
The physicsGAN discriminator compares these profiles with a feasible dataset ($\mathbf{x}$) prepared beforehand, as described in Sec.~\ref{sec:physicsGANperform}.
At the same time, the surrogate models take the twinGAN noise inputs and predict the aircraft state variables, which determine the constraint-based penalties.
During training, the optimizer updates the network weights ($\mathbf{W}_\mathbf{g}$ and $\mathbf{W}_\mathbf{d}$) using both the discriminator's probabilities ($p_\mathbf{g}$ and $p_\mathbf{x}$) and the accumulated penalty ($\lambda$).
In the end, the physicsGAN reliably produces valid $\mathbf{z}_t$ and ${T}_{\text{total}}$ pairs that give feasible control profiles for a range of flight conditions.

Adding the physicsGAN generator and the constraint penalties to the standard adversarial minimax framework gives the modified loss function below:
\begin{equation}
    \label{eq:physicsGAN}
    \min_{G_c} \max_{D_c} V\left(D_c,G_c\right)=E_{\mathbf{x} \sim p_{\text{data}}\left(\mathbf{x}\right)}\left[\log D_c\left(\mathbf{x},\mathbf{q}\right)\right]+E_{\mathbf{z} \sim p_\mathbf{z}\left(\mathbf{z}\right)}\left[\log \left(1-D_c\left(G_p\left(G_c\left(\mathbf{z},\mathbf{q}\right)\right),G_t\left(G_c\left(\mathbf{z},\mathbf{q}\right)\right),\mathbf{q}\right)\right)\right] + \lambda,
\end{equation}
where $V$ is the loss function, and $G_{c}$ and $D_{c}$ are the physicsGAN generator and discriminator, respectively.
$G_p$ and $G_t$ are the frozen twinGAN generators of the power and wing angle profiles.
In addition, $\mathbf{x}$ is drawn from the true feasible data distribution $P_{\text{data}}$, $\mathbf{z}$ is sampled from Uniform(0, 1) noise distributions ($P_z$), and $\lambda$ measures the penalty for constraint violations.
Note that the total penalty $\lambda$ sums the squared violations of the individual trajectories,
where $\lambda$ represents the acceleration constraint penalty for this work and is calculated as $(0.3-a_{max})^2$ if $a_{max} > 0.3 g$, where $g$ is the gravitational acceleration.

In terms of neural structure, the physicsGAN generator has five hidden layers of 128, 256, 512, 1,024, and 2,048 neurons. Each hidden layer is followed by a \texttt{LeakyReLU} activation ($\alpha=0.2$) and a batch normalization layer (momentum of 0.8), and the generator ends in a four-neuron output layer with a \texttt{sigmoid} activation. The physicsGAN discriminator, by contrast, has three hidden layers (1,024, 512, and 256 neurons) and ends in a single-neuron \texttt{sigmoid} output layer. Training lasts $3\times10^4$ epochs with a batch size of 400 and the \texttt{Adam} optimizer at a learning rate of $2\times10^{-4}$. Unlike standard GAN architectures, the physicsGAN offers a measurable performance metric called the coverage percentage. 
This metric is the proportion of strictly feasible designs in a random sample across the flight condition space, and it gives an explicit way to track convergence and to pick the best model iteration during training.

\section{Neural Envelope}
\label{sec:envelope}
\begin{figure}[htbp!]
    \centering
    \adjustbox{trim=3cm 0cm 4cm 0cm,clip=true}{ 
    \includegraphics[width=0.75\linewidth]{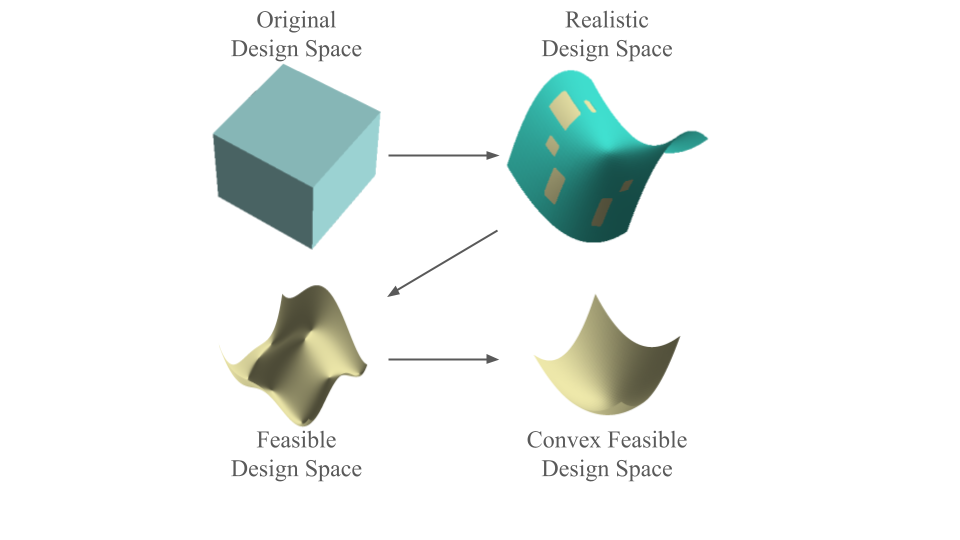}}
    \caption{Graphical representation of the desired design space transformation provided by the twinGAN, physicsGAN, and neural envelope.}
    \label{fig:DesignSpace}
\end{figure}

Figure~\ref{fig:DesignSpace} gives an illustrative overview of the expected design space transformation using twinGAN, physicsGAN, and neural envelope models. 
The cube on the top left is the original, constrained design space.
The twinGAN maps this space onto a realistic space (top right) of near-optimal takeoff control profiles, which naturally contains disjoint feasible (tan) and infeasible (blue) subregions.
Next, the physicsGAN explores the feasible space (bottom left), so that every physical constraint is inherently satisfied by construction.
Finally, the neural envelope increases the convexity of the space resulting in a unimodal, feasible space (bottom right).

The \texttt{ConvexHull} algorithm~\cite{SciPy} is a computational geometry method that identifies the exterior points of a point cloud.
It begins with the extreme points along each axis, which define an initial mesh.
For each facet of the mesh, the point farthest outside that facet is added as a new vertex, and points inside the mesh are discarded.
Coplanar or non-convex facets caused by floating-point error are merged to preserve convexity.
The algorithm returns the mesh as the vertices of each facet, together with the equation of the plane containing each facet.
The hull is a closed surface, and its lower portion is the convex envelope of the point cloud.
Selecting the facets whose normal vectors point downward delivers this convex envelope.
Each of the resulting planes defines an affine function, and the pointwise maximum of these functions is a convex, unimodal surface.

\begin{figure}[htbp!]
    \centering
    \includegraphics[width=0.75\textwidth]{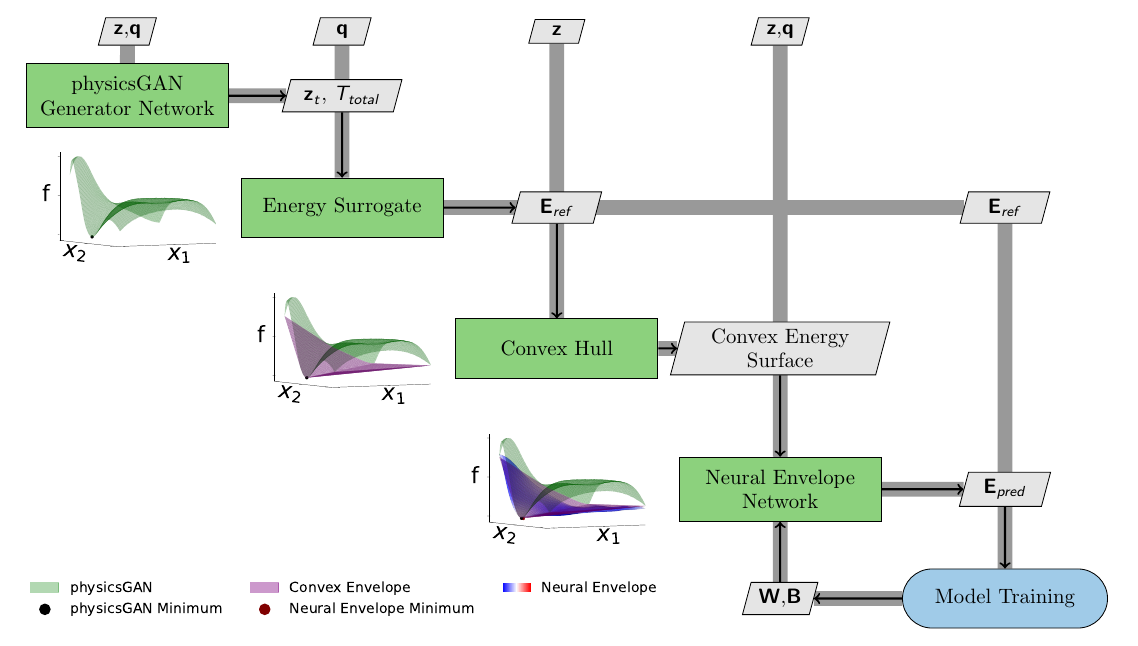}
    \caption{The neural envelope process diagram describes how the training data for the neural envelope is obtained and how the model is trained.
    }
    \label{fig:CE_process}
\end{figure}

The neural envelope process diagram (Fig.~\ref{fig:CE_process}) shows how the information and data flow within the framework.
The neural envelope in this work is a DNN trained on data from the convex envelopes of takeoff energy consumption within the physicsGAN variable space.  
The physicsGAN is conditioned on the flight conditions ($\eta$, $\mathbf{m}$), and the neural envelope is intended to learn the convex envelope over the flight-condition space instead of an individual flight condition.
To build the training data, the flight condition space is sampled uniformly, and a convex envelope is computed at each sampled flight condition with respect to the energy surrogate.
Each envelope is then evaluated on a 30 by 30 grid.
Together, these evaluations form the dataset on which the neural envelope is trained, enabling the neural envelope to represent the convex envelope of the physicsGAN at every flight condition.

\begin{figure}[htbp!]
    \centering
    \includegraphics[width=0.75\textwidth]{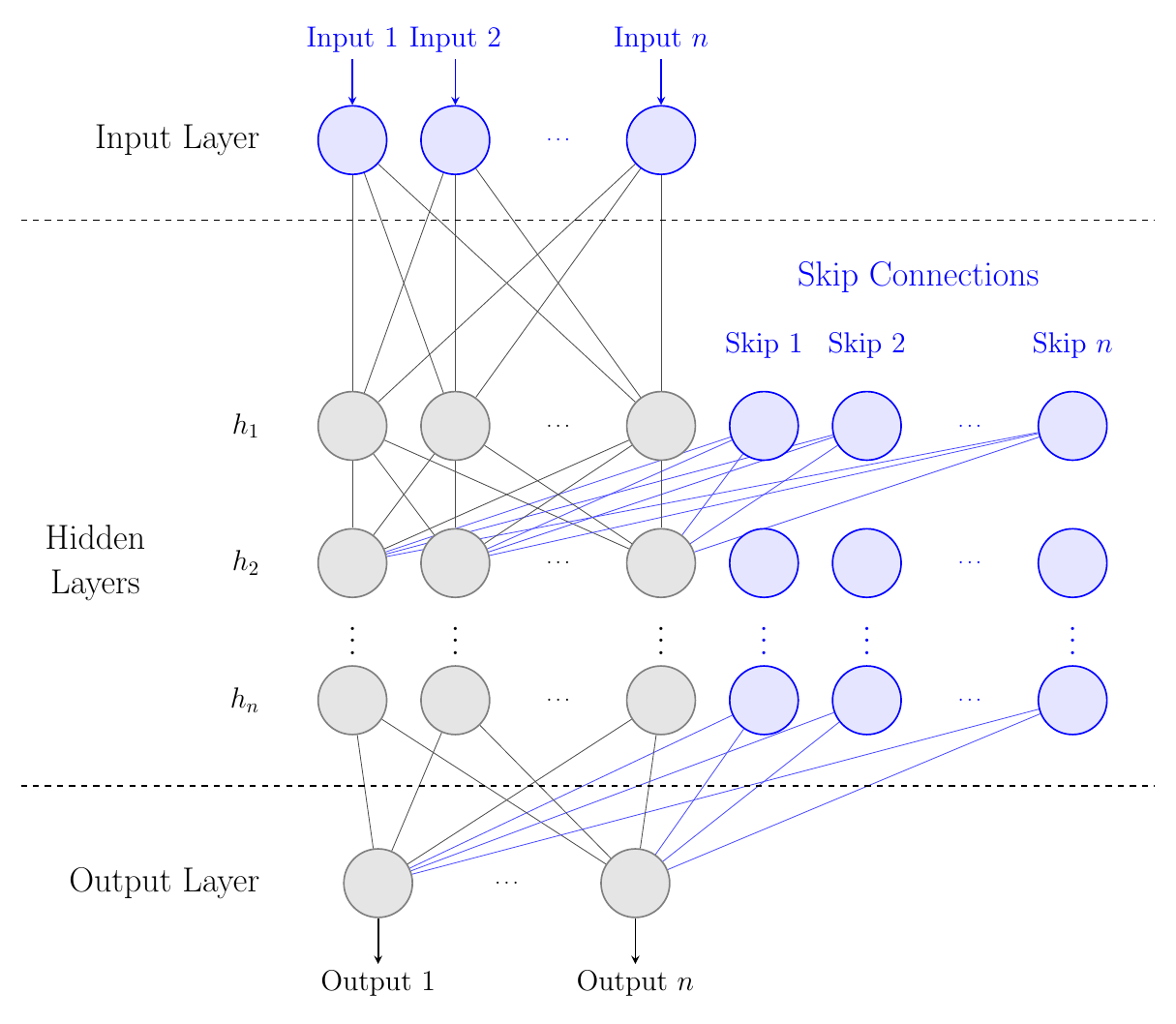}
    \caption{Deep neural network architecture with skip connections shown in blue. The skip connections pass the network inputs into the hidden and output layers.}
    \label{fig:DNN_skip}
\end{figure}

The neural envelope takes as inputs the physicsGAN noise variables ($\mathbf{z}$) and an arbitrary flight-condition label (${q}$).
Its architecture has six hidden layers of 128, 1028, 2056, 2056, 1028, and 64 neurons, respectively.
Each hidden layer uses a modified softplus activation function ($\phi$),
\begin{equation}
    \phi = \ln(1.0 + e^x) - \ln(2.0),
\end{equation}
which keeps the network smooth and differentiable while improving its representational capacity.
Skip connections (Fig.~\ref{fig:DNN_skip}) reinject the network inputs into each hidden layer.
The network outputs a single value, the energy ($E$) consumed during aircraft takeoff.
The energy consumption is log scaled to reduce its magnitude, which facilitates training, and then normalized to support the convex envelope operation.
The neural envelope is trained for 2,000 epochs with a batch size of 8,192 on a MSE loss function using the \texttt{Adam} optimizer and a learning rate of $1\times10^{-3}$.

\section{Results}
\label{sec:results}
This section presents how well the neural envelope performs on optimal takeoff trajectory design for the Airbus $A^3$ Vahana aircraft.
The neural envelope accuracy and optimization efficiency are examined across the flight condition space.
The neural envelope and physicsGAN optimization performance are compared as well.

\subsection{Problem Formulation}
\label{sec:probForm}
The $A^3$ Vahana's control profiles, \emph{i.e.}, the power delivered to the motors ($\mathbf{p}$), the wing angle relative to the vertical ($\boldsymbol{\theta}$), and the takeoff duration ($T_{\text{total}}$), are optimized to minimize the energy consumed during takeoff (Table~\ref{tab:Optimization Problem Statement}).
For the simulation-based MDAO, the power and wing angle profiles are parameterized by B-spline control points, whereas the twinGAN and physicsGAN encode those control points with noise variables.
Power is limited to [1, 311] kW, wing angle to [0, 135] degrees, and takeoff duration to [25, 35] seconds.
During takeoff, the aircraft's acceleration is constrained to ($a_{max}$).
To study how the physicsGAN performs, 
the power efficiency ($\eta$) and mass ($m$) are varied uniformly over the ranges [0.8, 1.0] and [616, 834] kg, respectively.
\begin{table}[htbp!]
    \caption{Problem formulation of the optimal takeoff trajectory design.}
    \label{tab:Optimization Problem Statement}
    \centering
    \begin{tabular}{p{2.5cm} p{3cm} p{6cm} p{2cm}}
        \hline
        \hline
        & \textbf{Function/Variable} & \textbf{Description} & \textbf{Quantity} \\
        \hline
        minimize & \textit{E} & Electrical energy & \\
        with respect to & $\textbf{x}$, $\textbf{z}_t$, or $\textbf{z}$ & B-spline control points, twinGAN variables, or physicsGAN variables & 40, 3, or 3 \\
        & $T_{\text{total}}$ & Total takeoff time & 1, 1, or N/A \\
        & & \textbf{Total design variables} & \textbf{41}, \textbf{4}, or \textbf{3} \\
        subject to & {$a_{max} \le 0.3$ g} & Acceleration constraint & 1 \\
        & & \textbf{Total constraint functions} & \textbf{1} \\
        \hline
        \hline
    \end{tabular}
\end{table}

\subsection{twinGAN and Surrogate Modeling}
Prior work~\cite{sisk2024surrogate} shows that twinGAN provides implicit dimensionality reduction by generating only realistic control profiles. 
Besides, twinGAN with three variables fits 1,099 optimal trajectories with 99\% accuracy for both power and wing angle generation.
The twinGAN explicitly reduces the input dimension from 41 to four (\emph{i.e.}, three twinGAN variables plus $T_{\text{total}}$).
The relative accuracy of the fitting optimizations is computed as
\begin{equation}
    \label{eq:14}
    \text{MAPRA} = \left(1 - \frac{1}{N} \sum_{{i}=1}^{N} \frac{\norm{\tilde{\mathbf{x}}_i - \mathbf{x}_i}_1}{\norm{\mathbf{x}_i}_1}\right)100\%,
\end{equation}
in which MAPRA is the mean absolute percent relative accuracy, $N$ = 1,099 is the number of training trajectories where each is discretized into 100 points, $\tilde{\mathbf{x}}_i$ are the generated discretized trajectory points, $\mathbf{x}_i$ are the reference discretized trajectory points, and $||\cdot||_1$ is the $\ell^1$ norm.

A feedforward DNN predicts the energy consumption, and LSTM surrogates predict the aircraft acceleration profile on which the surrogate-based MDAO relies.
Surrogate modeling is made easier by the implicit and explicit dimensionality reductions from the twinGAN.
In particular, the surrogate inputs are the twinGAN noise variables ($\mathbf{z_t}$), the flight conditions ($\mathbf{\eta}$, $\mathbf{m}$), and the takeoff duration ($T_{\text{total}}$).
The training, validation, and testing datasets contain 5,504, 1,376, and 2,573 simulation-based samples, respectively.
The surrogate models achieved 99.90\% and 99.87\% accuracy for the energy and acceleration surrogates, respectively.

\subsection{physicsGAN Performance}
\label{sec:physicsGANperform}
Training of the physicsGAN follows the procedure in Fig.~\ref{fig:ConGAN_XDSM}.
The generator and discriminator are trained on a batch size of 121 out of 10,601 feasible training samples.
These feasible samples originate from an 11 by 11 grid of points in the flight-condition space.
At each flight-condition point, the twinGAN generates realistic control profiles from randomly sampled noise variables ($\mathbf{z_t}$), paired with flight durations ($\mathbf{T}_{\text{total}}$).
The acceleration surrogate model checks whether the twinGAN-generated profiles are feasible.
In the end, 100 feasible profiles are acquired at each flight condition point.

The feasibility coverage percentage, meaning the portion of trajectories that satisfy the constraints in the transformed feasible design space, serves as the metric of physicsGAN performance.
To compute the coverage, 40,000 samples are randomly drawn from the flight-condition and noise spaces.
These samples are passed through the physicsGAN to get proposed trajectories and then the surrogate models verify the feasibility.
The physicsGAN model that reaches the highest coverage percentage during training is saved for further applications.

Besides the feasibility converage, the physicsGAN's variability is verified through fitting optimizations on 1,099 reference trajectories (Table~\ref{tab:pGAN}), which guarantees that actual optimal designs are included within the reduced space.
Each fitting optimization finds the optimal physicsGAN variables that reconstruct a given reference optimal profile.
Once the fitting optimizations finish, the mean relative accuracy is evaluated using Eqn.~\ref{eq:14}, where $N$ = 1,099 is the number of reference trajectories where each is discretized into 100 points, $\mathbf{x}_i$ denotes the reference discretized profile points, and $\tilde{\mathbf{x}}_i$ denotes the generated discretized profile points.
\begin{table}[htbp!]
    \centering
    \caption{Summary of physicsGAN performance.}
    \begin{tabular}{c c c c c}
        \hline
        \hline
        Model & Latent Variables & Coverage Percentage & Power Fitting Accuracy & Wing Angle Fitting Accuracy \\
        \hline
        physicsGAN & 2 & 99.20\% & 99.51\% & 99.61\% \\
        \hline
        \hline
    \end{tabular}
    \label{tab:pGAN}
\end{table}

\subsection{Neural Envelope Performance}
\label{sec:NEperform}
The neural envelope is trained with 400 training, 100 validation, and 200 testing flight conditions.
At each of those flight conditions, a 30 by 30 grid is sampled within the physicsGAN variable space to construct the reference convex envelope using \texttt{ConvexHull}.
This provides final dataset sizes of 360,000, 90,000, and 180,000, respectively.
Each dataset is shuffled before training the neural envelope.

The model with the lowest validation loss is saved to use for the neural-envelope-enabled surrogate-based optimization, resulting in high accuracy (Table~\ref{tab:CE}).
The envelope MAPRA compares the neural envelope predicted surface against the reference convex envelope, while the global minimum MAPRA and global minimum location normalized root mean squared error (NRMSE) measure how well the neural envelope matches the global minimum of the physicsGAN.
The global minimum MAPRA compares the optimum energy consumption of the physicsGAN and neural envelope.
The global minimum NRMSE measures the distance in the $\mathbf{z}$ space between the minimums of the physicsGAN and neural envelope.
\begin{table}[htbp!]
    \centering
    \caption{Summary of neural envelope accuracies.}
    \begin{tabular}{c c c c}
        \hline
        \hline
        Model & Envelope MAPRA & Global Minimum Energy MAPRA & Global Minimum Location NRMSE \\
        \hline
        Neural Envelope & 99.91\% & 99.84\% & 0.018567 \\
        \hline
        \hline
    \end{tabular}
    \label{tab:CE}
\end{table}

\begin{figure}[htbp!]
    \centering
    \includegraphics[width=0.75\linewidth]{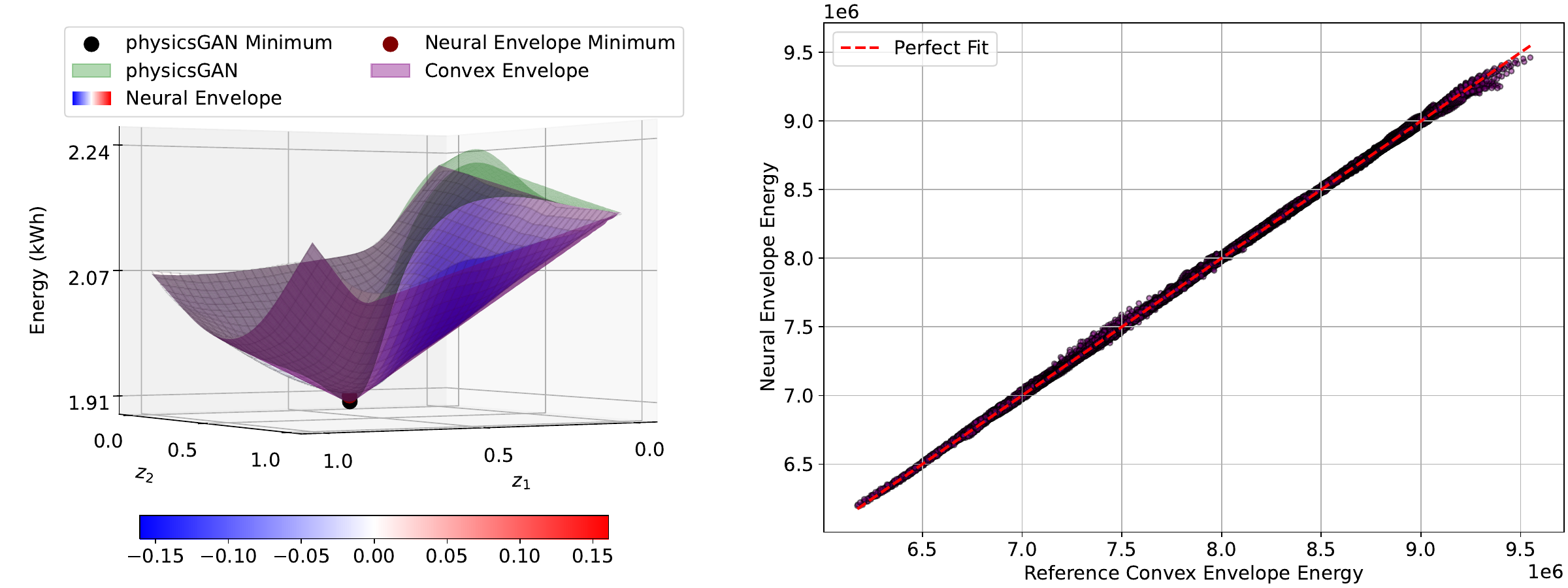}
    \caption{
    Comparison of the physicsGAN, convex envelope, and neural envelope at Point~1~($\eta=98.01\%$, $m=669.24$~kg).
    The neural envelope achieves an accuracy of 99.88\% and matches the global optimum with 99.61\% accuracy.
    }
    \label{fig:envelope1}
\end{figure}

\begin{figure}[htbp!]
    \centering
    \includegraphics[width=0.75\linewidth]{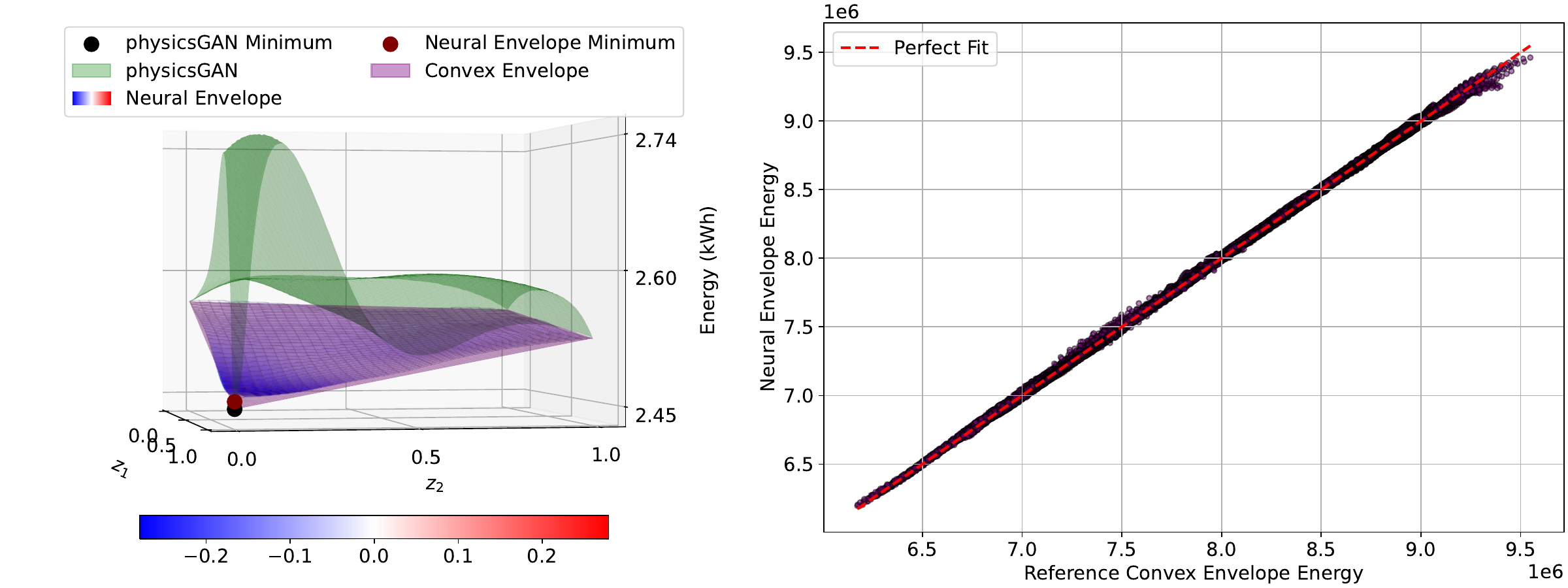}
    \caption{
    Comparison of the physicsGAN, convex envelope, and neural envelope at Point~2~($\eta=80.61\%$, $m=829.07$~kg).
    The neural envelope achieves an accuracy of 99.78\% and matches the global optimum with 99.68\% accuracy.
    }
    \label{fig:envelope2}
\end{figure}

\begin{figure}[htbp!]
    \centering
    \includegraphics[width=0.75\linewidth]{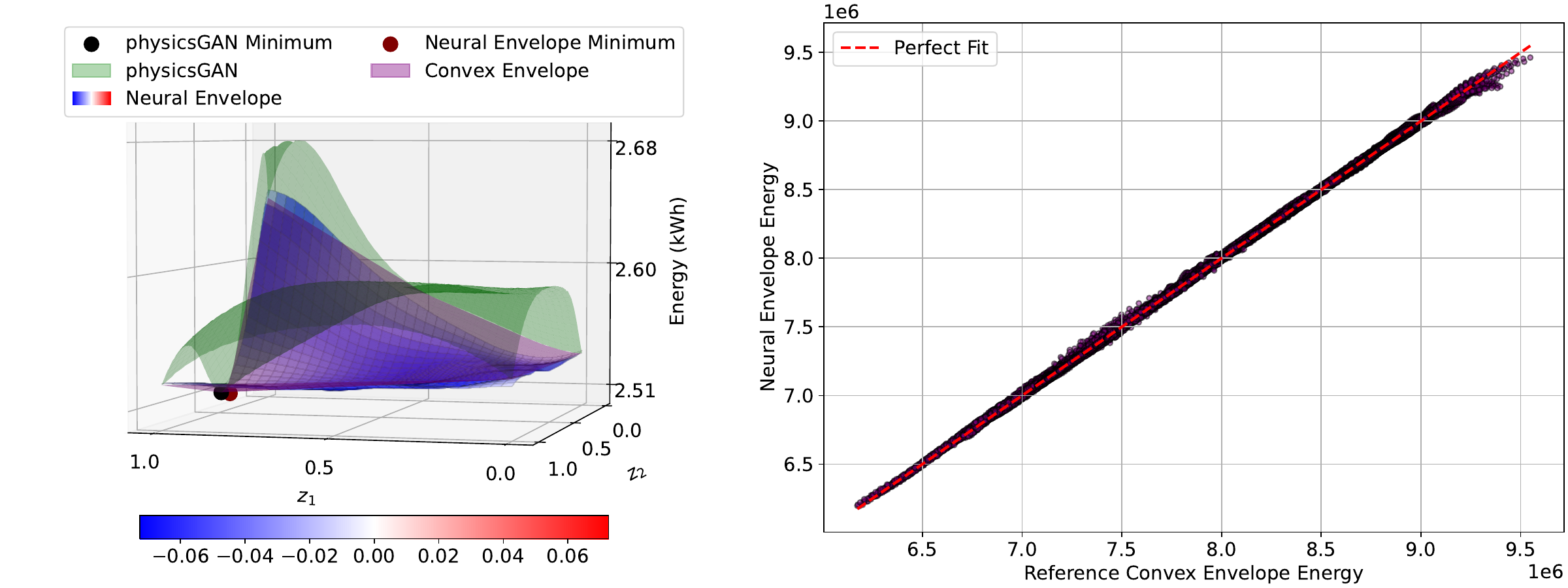}
    \caption{
    Comparison of the physicsGAN, convex envelope, and neural envelope at Point~3~($\eta=81.52\%$, $m=826.12$~kg).
    The neural envelope achieves an accuracy of 99.88\% and matches the global optimum with 99.97\% accuracy.
    }
    \label{fig:envelope3}
\end{figure}

Three arbitrary flight conditions, Point~1~($\eta=98.01\%$, $m=669.24$ kg), Point~2~($\eta=80.61\%$, $m=829.07$ kg), and Point~3~($\eta=81.52\%$, $m=826.12$ kg),  with clear multimodality are chosen to visualize the physicsGAN, convex envelope, and neural envelope surfaces (Figs.~\ref{fig:envelope1}~--~\ref{fig:envelope3}).
Results show that the physicsGAN has a multimodal surface and the neural envelope consistently produces a unimodal, near-convex surface.
This smooth, unimodal surface enables efficient gradient-based optimization as shown in Sec.~\ref{sec:opt}.

\subsection{Neural Envelope-enabled Takeoff Trajectory Optimizations}
\label{sec:opt}
To achieve a thorough comparison, the trajectory optimization is performed with three optimizers for both the physicsGAN and neural envelope in this section (Table~\ref{tab:test_opt_results}).
A parametric study of global optimizers led to the choice of the \texttt{Differential Evolution}~\cite{storn1997differential} (DE) optimizer from \texttt{SciPy}~\cite{SciPy}, which uses a population of 100 and a tolerance of $1\times10^{-4}$.
Additionally, two gradient-based methods, \texttt{SLSQP} and \texttt{L-BFGS-B} from \texttt{SciPy} with a tolerance of $1\times10^{-4}$, are chosen.
To increase the probability of the gradient-based optimizers finding the global minimum, they are each run 20 times at each flight condition from randomized starting locations.
The gradient-based global minimum is selected from those 20 runs, and the corresponding optimization duration is recorded.
Meanwhile only a single optimization is done with DE and the total optimization duration is recorded.
\begin{table}[htbp!]
    \centering
    \caption{Summary of average testing data optimization durations.}
    \begin{tabular}{l c c c}
        \hline
        \hline
        Model & DE Duration (s) & SLSQP Duration (s) & L-BFGS-B Duration (s) \\
        \hline
        physicsGAN & 6.321 & 0.177 & 0.103 \\
        Neural Envelope & 2.815 & 0.071 & 0.039 \\
        \hline
        \hline
    \end{tabular}
    \label{tab:test_opt_results}
\end{table}

Table~\ref{tab:test_opt_results} presents that the neural-envelope-enabled optimization is consistently 2--3 times faster than the physicsGAN optimization.
The neural envelope is faster for the following three reasons.
First, the neural envelope involves only one network for evaluations, while the physicsGAN requires two networks for evaluations.
Second, the neural envelope utilizes smooth activation functions to provide continuous derivatives for the gradient-based optimization.
Third, the neural envelope, being unimodal, allows the gradient-based optimizer to find the global optima with only a single starting point.
Thus, the neural envelope reduces the cumulative physicsGAN global optimization duration from 6.32~s with DE or 2.06~s with 20 L-BFGS-B runs down to 0.04~s.



\section{Conclusions}
\label{sec:conclusion}
The excessive energy consumption of electric vertical takeoff and landing (eVTOL) aircraft is still a key obstacle for broad deployment. 
Surrogate-based design optimization is applied to lower takeoff energy requirements while mitigating the heavy computational cost of conventional approaches.
Especially, dimensionality reduction promotes the surrogate modeling in large-scale engineering applications. 
For instance, the data-driven twin-generator generative adversarial networks (twinGAN) in this work reduces dimensionality both implicitly and explicitly.
Still, practical problems often involve intricate nonlinear constraints, which make design optimization harder by obstructing the discovery of feasible regions.
In addition, neural networks are not guaranteed to be unimodal and require global optimization over gradient-based optimization.

To address this difficulty, this study proposes the combination of physics-constrained generative adversarial networks (physicsGAN) and neural envelope.
The physicsGAN discovers the feasible design space and enables unconstrained optimization, while the neural envelope transforms the multimodal feasible physicsGAN space into a unimodal, near-convex feasible space. 
Specifically, the physicsGAN exploits a sub-set distribution of twinGAN variables which, through the energy surrogate, produce the feasible control profiles.
Then the reference convex envelope with respect to physicsGAN variable space is generated for each flight condition of interest.
This enables the neural envelope to approximate the convex envelope for the entire flight-condition space and enable efficient gradient-based optimization.

The coupling between physicsGAN and neural envenlope is demonstrated on optimum takeoff trajectory design for the Airbus $A^3$ Vahana eVTOL aircraft, with minimum energy consumption as the objective. 
Over the entire mass-efficiency condition space, physicsGAN manages to discover a feasible space on which the neural envelope learns the convex envelope with 99.91\% accuracy.
Comparing the global minimum of the neural envelope to the physicsGAN, the neural envelope achieves 99.84\% energy accuracy and the location of the global minimums had an NRMSE of less than 0.02.
These results ensure the neural envelope is accurately replicating the global minimum of the physicsGAN while using a single unimodal network.

Surrogate optimization driven by the neural envelope is evaluated across a variety of mass and efficiency configurations. 
Results show that unimodal design space permits rapid gradient-based optimization over global optimization.  
The neural envelope-based optimization finishes in 0.04 seconds on average, which is 53 times faster than the physicsGAN-enabled gradient-based optimization and 162 times faster than the physicsGAN-enabled global optimization. 
The neural envelope approach, therefore, paves a new avenue for real-time UAM trajectory optimization under intricate nonlinear constraints.

Future work will study how the neural envelope scales to higher dimension input spaces, as well as learning from physicsGAN models that handle more constraints.
In addition, the neural envelope will be expanded to produce a rigorously convex space, rather than a near-convex space.

\newpage
\bibliography{phlai,sample,Aviation}

@article{sisk2026physics,
  title={Physics-constrained generative adversarial networks for rapid optimal takeoff trajectory designs},
  author={Sisk, Samuel and Du, Xiaosong},
  journal={Structural and Multidisciplinary Optimization},
  volume={69},
  number={6},
  pages={160},
  year={2026},
  publisher={Springer}
}

@article{yang2021optimization,
  title={Optimization-based control using input convex neural networks},
  author={Yang, Shu and Bequette, B Wayne},
  journal={Computers \& Chemical Engineering},
  volume={144},
  pages={107143},
  year={2021},
  publisher={Elsevier}
}

@inproceedings{ahmadi2013piecewise,
  title={Piecewise linear approximation of generators cost functions using max-affine functions},
  author={Ahmadi, Hamed and Mart{\'\i}, Jos{\'e} R and Moshref, Ali},
  booktitle={2013 IEEE Power \& Energy Society General Meeting},
  pages={1--5},
  year={2013},
  organization={IEEE}
}

@article{pham2009convex,
  title={Convex hull discretization approach to the global optimization of pooling problems},
  author={Pham, Viet and Laird, Carl and El-Halwagi, Mahmoud},
  journal={Industrial \& Engineering Chemistry Research},
  volume={48},
  number={4},
  pages={1973--1979},
  year={2009},
  publisher={ACS Publications}
}

@Article{Du2021a,
    author      = {Xiaosong Du and Ping He and Joaquim R. R. A. Martins},
    title       = {Rapid Airfoil Design Optimization via Neural Networks-based Parameterization and Surrogate Modeling},
    doi         = {10.1016/j.ast.2021.106701},
    journal     = {Aerospace Science and Technology},
    month       = {June},
    pages       = {106701},
    volume      = {113},
    year        = {2021}
}

@Article{Gill2002a,
    author      = {Philip E. Gill and Walter Murray and Michael A. Saunders},
    title       = {{SNOPT}: {A}n {SQP} algorithm for large-scale constrained optimization},
    doi         = {10.1137/S1052623499350013},
    journal     = {SIAM Journal of Optimization},
    number      = {4},
    pages       = {979--1006},
    volume      = {12},
    year        = {2002}
}

@InCollection{Goodfellow2014,
    author      = {Goodfellow, Ian and Pouget-Abadie, Jean and Mirza, Mehdi and Xu, Bing and Warde-Farley, David and Ozair, Sherjil and Courville, Aaron and Bengio, Yoshua},
    title       = {Generative Adversarial Nets},
    booktitle   = {Advances in Neural Information Processing Systems 27},
    editor      = {Z. Ghahramani and M. Welling and C. Cortes and N. D. Lawrence and K. Q. Weinberger},
    pages       = {2672--2680},
    publisher   = {Curran Associates, Inc.},
    year        = {2014}
}

@Article{Karmarkar1984,
    author      = {N. Karmarkar},
    title       = {A new polynomial-time algorithm for linear programming},
    doi         = {10.1007/bf02579150},
    journal     = {Combinatorica},
    month       = {dec},
    number      = {4},
    pages       = {373--395},
    publisher   = {Springer Nature},
    volume      = {4},
    year        = 1984
}

@InCollection{Leifsson2011surrogate,
    author      = {Slawomir Koziel and David Echeverr{\'{\i}}a Ciaurri and Leifur Leifsson},
    title       = {Surrogate-Based Methods},
    booktitle   = {Computational Optimization, Methods and Algorithms},
    doi         = {10.1007/978-3-642-20859-1_3},
    pages       = {33--59},
    publisher   = {Springer Berlin Heidelberg},
    url         = {https://doi.org/10.1007/978-3-642-20859-1_3},
    year        = {2011}
}

@Article{Peherstorfer2018a,
    author      = {Peherstorfer, Benjamin. and Willcox, Karen. and Gunzburger, Max},
    title       = {Survey of Multifidelity Methods in Uncertainty Propagation, Inference, and Optimization},
    doi         = {10.1137/16M1082469},
    issn        = {0036-1445},
    journal     = {SIAM Review},
    month       = {January},
    number      = {3},
    pages       = {550--591},
    volume      = {60},
    year        = {2018}
}

@Article{Sherstinsky2020,
    author      = {Sherstinsky, Alex},
    title       = {Fundamentals of Recurrent Neural Network ({RNN}) and Long Short-Term Memory ({LSTM}) network},
    doi         = {10.1016/j.physd.2019.132306},
    issn        = {0167-2789},
    journal     = {Physica D: Nonlinear Phenomena},
    month       = {Mar},
    pages       = {132306},
    publisher   = {Elsevier BV},
    volume      = {404},
    year        = {2020}
}

@Misc{tensorflow2015-whitepaper,
    author      = {Abadi, Mart{\'{i}}n and Agarwal, Ashish and Barham, Paul and Brevdo, Eugene and Chen, Zhifeng and Citro, Craig and Corrado, Greg S. and Davis, Andy and Dean, Jeffrey and Devin, Matthieu and Ghemawat, Sanjay and Goodfellow, Ian and Harp, Andrew and Irving, Geoffrey and Isard, Michael and Jia, Yangqing and Jozefowicz, Rafal and Kaiser, Lukasz and Kudlur, Manjunath and Levenberg, Josh and Mane, Dan and Monga, Rajat and Moore, Sherry and Murray, Derek and Olah, Chris and Schuster, Mike and Shlens, Jonathon and Steiner, Benoit and Sutskever, Ilya and Talwar, Kunal and Tucker, Paul and Vanhoucke, Vincent and Vasudevan, Vijay and Viegas, Fernanda and Vinyals, Oriol and Warden, Pete and Wattenberg, Martin and Wicke, Martin and Yu, Yuan and Zheng, Xiaoqiang},
    title       = {{TensorFlow}: Large-Scale Machine Learning on Heterogeneous Distributed Systems},
    isbn        = {0010-0277},
    issn        = {0270-6474},
    url         = {http://arxiv.org/abs/1603.04467},
    year        = {2016}
}

@article{straubinger2020overview,
  title={An overview of current research and developments in urban air mobility--Setting the scene for UAM introduction},
  author={Straubinger, Anna and Rothfeld, Raoul and Shamiyeh, Michael and B{\"u}chter, Kai-Daniel and Kaiser, Jochen and Pl{\"o}tner, Kay Olaf},
  journal={Journal of Air Transport Management},
  volume={87},
  pages={101852},
  year={2020},
  publisher={Elsevier}
}

@article{chauhan2020tilt,
  title={Tilt-wing eVTOL takeoff trajectory optimization},
  author={Chauhan, Shamsheer S and Martins, Joaquim RRA},
  journal={Journal of Aircraft},
  volume={57},
  number={1},
  pages={93--112},
  year={2020},
  publisher={American Institute of Aeronautics and Astronautics}
}

@inproceedings{falck2017trajectory,
  title={Trajectory optimization of electric aircraft subject to subsystem thermal constraints},
  author={Falck, Robert D and Chin, Jeffrey and Schnulo, Sydney L and Burt, Jonathan M and Gray, Justin S},
  booktitle={18th AIAA/ISSMO Multidisciplinary Analysis and Optimization Conference},
  pages={4002},
  year={2017}
}

@inproceedings{alba2017surrogate,
  title={A surrogate-based multi-disciplinary design optimization framework exploiting wing-propeller interaction},
  author={Alba, Christian and Elham, Ali and German, Brian and Veldhuis, Leo L},
  booktitle={18th AIAA/ISSMO Multidisciplinary Analysis and Optimization Conference},
  pages={4329},
  year={2017}
}

@article{swischuk2020learning,
  title={Learning physics-based reduced-order models for a single-injector combustion process},
  author={Swischuk, Renee and Kramer, Boris and Huang, Cheng and Willcox, Karen},
  journal={AIAA Journal},
  volume={58},
  number={6},
  pages={2658--2672},
  year={2020},
  publisher={American Institute of Aeronautics and Astronautics}
}

@Article{SciPy,
    author      = {SciPy},
    editor      = {Virtanen, Pauli and Gommers, Ralf and Oliphant, Travis E and Haberland, Matt and Reddy, Tyler and Cournapeau, David and Burovski, Evgeni and Peterson, Pearu and Weckesser, Warren and Bright, Jonathan and van der Walt, St{\'{e}}fan J and Brett, Matthew and Wilson, Joshua and Millman, K Jarrod and Mayorov, Nikolay and Nelson, Andrew R J and Jones, Eric and Kern, Robert and Larson, Eric and Carey, C J and Polat, $\backslash$.Ilhan and Feng, Yu and Moore, Eric W and VanderPlas, Jake and Laxalde, Denis and Perktold, Josef and Cimrman, Robert and Henriksen, Ian and Quintero, E A and Harris, Charles R and Archibald, Anne M and Ribeiro, Ant{\^{o}}nio H and Pedregosa, Fabian and van Mulbregt, Paul and {SciPy 1.0 Contributors}},
    title       = {{SciPy} 1.0: Fundamental Algorithms for Scientific Computing in Python},
    doi         = {10.1038/s41592-019-0686-2},
    journal     = {Nature Methods},
    pages       = {261--272},
    volume      = {17},
    year        = {2020}
}

@article{SNOPT,
author = {Gill, Philip E. and Murray, Walter and Saunders, Michael A.},
title = {SNOPT: An SQP Algorithm for Large-Scale Constrained Optimization},
year = {2005},
issue_date = {2005},
publisher = {Society for Industrial and Applied Mathematics},
address = {USA},
volume = {47},
number = {1},
issn = {0036-1445},
url = {https://doi.org/10.1137/S0036144504446096},
doi = {10.1137/S0036144504446096},
journal = {SIAM Rev.},
month = {jan},
pages = {99–131},
numpages = {33}
}

@misc{kingma2017adam,
      title={Adam: A Method for Stochastic Optimization}, 
      author={Diederik P. Kingma and Jimmy Ba},
      year={2017},
      eprint={1412.6980},
      archivePrefix={arXiv},
      primaryClass={cs.LG}
}

@misc{texas2019transportation,
  title={Transportation Institute. Urban mobility report},
  author={Texas, AM},
  year={2019}
}

@inproceedings{courtin2018feasibility,
  title={Feasibility study of short takeoff and landing urban air mobility vehicles using geometric programming},
  author={Courtin, Christopher and Burton, Michael J and Yu, Alison and Butler, Patrick and Vascik, Parker D and Hansman, R John},
  booktitle={2018 Aviation Technology, Integration, and Operations Conference},
  pages={4151},
  year={2018}
}

@phdthesis{vascik2017systems,
  title={Systems-level analysis of On Demand Mobility for aviation},
  author={Vascik, Parker D},
  year={2017},
  school={Massachusetts Institute of Technology}
}

@inproceedings{duffy2017study,
  title={A study in reducing the cost of vertical flight with electric propulsion},
  author={Duffy, Michael J and Wakayama, Sean R and Hupp, Ryan},
  booktitle={17th AIAA Aviation Technology, Integration, and Operations Conference},
  pages={3442},
  year={2017}
}

@techreport{edwards2020evtol,
  title={eVTOL passenger acceptance},
  author={Edwards, Thomas and Price, George},
  year={2020},
  institution={NASA}
}

@article{gao2024noise,
  title={Noise-Aware and Equitable Urban Air Traffic Management: An Optimization Approach},
  author={Gao, Zhenyu and Yu, Yue and Wei, Qinshuang and Topcu, Ufuk and Clarke, John-Paul},
  journal={arXiv preprint arXiv:2401.00806},
  year={2024}
}

@inproceedings{harrison2019using,
  title={Using multi-physics system simulation to predict battery pack thermal performance and risk of Thermal runaway during eVTOL aircraft operations},
  author={Harrison, Jonathan and Charles, Devin and Zenker, Jon and Frank, Evan},
  booktitle={2019 AIAA/IEEE Electric Aircraft Technologies Symposium (EATS)},
  pages={1--13},
  year={2019},
  organization={IEEE}
}

@book{mdobook,
author = {Joaquim R. R. A. Martins and Andrew Ning},
title = {Engineering Design Optimization},
doi = {10.1017/9781108980647},
isbn = {9781108833417},
month = {January},
publisher = {Cambridge University Press},
address = {Cambridge, UK},
url = {https://mdobook.github.io},
year = {2022}
}

@article{lazzara2022surrogate,
  title={Surrogate modelling for an aircraft dynamic landing loads simulation using an LSTM AutoEncoder-based dimensionality reduction approach},
  author={Lazzara, Michele and Chevalier, Max and Colombo, Michele and Garcia, Jasone Garay and Lapeyre, Corentin and Teste, Olivier},
  journal={Aerospace Science and Technology},
  volume={126},
  pages={107629},
  year={2022},
  publisher={Elsevier}
}

@inproceedings{andre2019robust,
  title={Robust environmental life cycle assessment of electric VTOL concepts for urban air mobility},
  author={Andr{\'e}, Nicolas and Hajek, Manfred},
  booktitle={AIAA aviation 2019 forum},
  pages={3473},
  year={2019}
}

@inproceedings{vascik2019development,
  title={Development of vertiport capacity envelopes and analysis of their sensitivity to topological and operational factors},
  author={Vascik, Parker D and Hansman, R John},
  booktitle={AIAA Scitech 2019 Forum},
  pages={0526},
  year={2019}
}

@article{hochreiter1997long,
  title={Long short-term memory},
  author={Hochreiter, Sepp and Schmidhuber, J{\"u}rgen},
  journal={Neural computation},
  volume={9},
  number={8},
  pages={1735--1780},
  year={1997},
  publisher={MIT press}
}

@article{sisk2024surrogate,
  title={Surrogate-Based Multidisciplinary Optimization for the Takeoff Trajectory Design of Electric Drones},
  author={Sisk, Samuel and Du, Xiaosong},
  journal={Processes},
  volume={12},
  number={9},
  pages={1864},
  year={2024},
  publisher={MDPI}
}

@techreport{goyal2018urban,
  title={Urban air mobility (UAM) market study},
  author={Goyal, Rohit and Reiche, Colleen and Fernando, Chris and Serrao, Jacquie and Kimmel, Shawn and Cohen, Adam and Shaheen, Susan},
  year={2018},
  institution={NASA}
}

@article{dan2020generative,
  title={Generative adversarial networks (GAN) based efficient sampling of chemical composition space for inverse design of inorganic materials},
  author={Dan, Yabo and Zhao, Yong and Li, Xiang and Li, Shaobo and Hu, Ming and Hu, Jianjun},
  journal={npj Computational Materials},
  volume={6},
  number={1},
  pages={84},
  year={2020},
  publisher={Nature Publishing Group UK London}
}

@article{yang2021challenges,
  title={Challenges and key requirements of batteries for electric vertical takeoff and landing aircraft},
  author={Yang, Xiao-Guang and Liu, Teng and Ge, Shanhai and Rountree, Eric and Wang, Chao-Yang},
  journal={Joule},
  volume={5},
  number={7},
  pages={1644--1659},
  year={2021},
  publisher={Elsevier}
}

@article{sripad2021promise,
  title={The promise of energy-efficient battery-powered urban aircraft},
  author={Sripad, Shashank and Viswanathan, Venkatasubramanian},
  journal={Proceedings of the National Academy of Sciences},
  volume={118},
  number={45},
  pages={e2111164118},
  year={2021},
  publisher={National Acad Sciences}
}

@article{mlaso,
  title={Machine learning in aerodynamic shape optimization},
  author={Li, Jichao and Du, Xiaosong and Martins, Joaquim RRA},
  journal={Progress in Aerospace Sciences},
  volume={134},
  pages={100849},
  year={2022},
  publisher={Elsevier}
}

@article{yang2020physics,
  title={Physics-informed generative adversarial networks for stochastic differential equations},
  author={Yang, Liu and Zhang, Dongkun and Karniadakis, George Em},
  journal={SIAM Journal on Scientific Computing},
  volume={42},
  number={1},
  pages={A292--A317},
  year={2020},
  publisher={SIAM}
}

@article{yu2024physics,
  title={Physics-guided generative adversarial network for probabilistic structural system identification},
  author={Yu, Yang and Liu, Yongming},
  journal={Expert Systems with Applications},
  volume={239},
  pages={122339},
  year={2024},
  publisher={Elsevier}
}

@article{di2020efficient,
  title={Efficient generation of structured objects with constrained adversarial networks},
  author={Di Liello, Luca and Ardino, Pierfrancesco and Gobbi, Jacopo and Morettin, Paolo and Teso, Stefano and Passerini, Andrea},
  journal={Advances in neural information processing systems},
  volume={33},
  pages={14663--14674},
  year={2020}
}

@article{zhang2024high,
  title={High-fidelity aero-acoustic evaluations of a heavy-lift eVTOL in hover},
  author={Zhang, Tao and Barakos, George N and Filippone, Antonio and others},
  journal={Journal of Sound and Vibration},
  volume={584},
  pages={118453},
  year={2024},
  publisher={Elsevier}
}

@misc{omalley2019kerastuner,
    title        = {KerasTuner},
    author       = {O'Malley, Tom and Bursztein, Elie and Long, James and Chollet, Fran\c{c}ois and Jin, Haifeng and Invernizzi, Luca and others},
    year         = 2019,
    howpublished = {\url{https://github.com/keras-team/keras-tuner}}
}

@article{storn1997differential,
  title={Differential evolution--a simple and efficient heuristic for global optimization over continuous spaces},
  author={Storn, Rainer and Price, Kenneth},
  journal={Journal of global optimization},
  volume={11},
  pages={341--359},
  year={1997},
  publisher={Springer}
}

@misc{Joby, title={Joby Aviation}, url={https://www.jobyaviation.com/}, journal={Joby},year=2025}

@misc{EHang, title={EHang}, url={https://www.ehang.com/}, journal={EHang}, year=2025}

@inproceedings{power2023sampling,
  title={Sampling constrained trajectories using composable diffusion models},
  author={Power, Thomas and Soltani-Zarrin, Rana and Iba, Soshi and Berenson, Dmitry},
  booktitle={IROS 2023 Workshop on Differentiable Probabilistic Robotics: Emerging Perspectives on Robot Learning},
  pages={n.pag.},
  year={2023},
  url={https://openreview.net/pdf?id=UAylEpIMNE}
}

@inproceedings{zhu2024controltraj,
  title={Controltraj: Controllable trajectory generation with topology-constrained diffusion model},
  author={Zhu, Yuanshao and Yu, James Jianqiao and Zhao, Xiangyu and Liu, Qidong and Ye, Yongchao and Chen, Wei and Zhang, Zijian and Wei, Xuetao and Liang, Yuxuan},
  booktitle={Proceedings of the 30th ACM SIGKDD Conference on Knowledge Discovery and Data Mining},
  pages={4676--4687},
  year={2024}
}

@inproceedings{guffanti2024transformers,
  title={Transformers for trajectory optimization with application to spacecraft rendezvous},
  author={Guffanti, Tommaso and Gammelli, Daniele and D’Amico, Simone and Pavone, Marco},
  booktitle={2024 IEEE Aerospace Conference},
  pages={1--13},
  year={2024},
  organization={IEEE}
}

@inproceedings{johnson2024zero,
  title={Zero-Shot Constrained Motion Planning Transformers Using Learned Sampling Dictionaries},
  author={Johnson, Jacob J and Qureshi, Ahmed H and Yip, Michael C},
  booktitle={2024 IEEE International Conference on Robotics and Automation (ICRA)},
  pages={14363--14369},
  year={2024},
  organization={IEEE}
}

@article{Bird2021,
title = {Dimensionality-reduction-based surrogate models for real-time design space exploration of a jet engine compressor blade},
journal = {Aerospace Science and Technology},
volume = {118},
pages = {107077},
year = {2021},
issn = {1270-9638},
doi = {https://doi.org/10.1016/j.ast.2021.107077},
author = {Gregory D. Bird and Steven E. Gorrell and John L. Salmon}
}

@article{Hou2022,
  title={Dimensionality Reduction in Surrogate Modeling: A Review of Combined Methods},
  author={Hou, Chun Kit Jeffery and Behdinan, Kamran},
  journal={Data Science and Engineering},
  volume={7},
  number={4},
  pages={402-427},
  year={2022},
  doi={10.1007/s41019-022-00193-5}
}

@incollection{werbos1982applications,
  author    = {Werbos, Paul J.},
  title     = {Applications of advances in nonlinear sensitivity analysis},
  booktitle = {System Modeling and Optimization},
  series    = {Lecture Notes in Control and Information Sciences},
  volume    = {38},
  editor    = {Drenick, R. F. and Kozin, F.},
  pages     = {762--770},
  year      = {1982},
  publisher = {Springer-Verlag},
  address   = {Berlin, Heidelberg},
  doi       = {10.1007/BFB0006173}
}

@Article{du2026generative,
AUTHOR = {Du, Xiaosong},
title = {Generative Artificial Intelligence in Aircraft Design Optimization},
JOURNAL = {Processes},
VOLUME = {14},
YEAR = {2026},
NUMBER = {4},
ARTICLE-NUMBER = {719},
ISSN = {2227-9717},
DOI = {10.3390/pr14040719}
}

@article{OLEARYROSEBERRY2022115730,
title = {Learning high-dimensional parametric maps via reduced basis adaptive residual networks},
journal = {Computer Methods in Applied Mechanics and Engineering},
volume = {402},
pages = {115730},
year = {2022},
note = {A Special Issue in Honor of the Lifetime Achievements of J. Tinsley Oden},
issn = {0045-7825},
doi = {https://doi.org/10.1016/j.cma.2022.115730},
url = {https://www.sciencedirect.com/science/article/pii/S0045782522006855},
author = {Thomas O’Leary-Roseberry and Xiaosong Du and Anirban Chaudhuri and Joaquim R.R.A. Martins and Karen Willcox and Omar Ghattas}
}

\end{document}